\documentclass{article} 
\usepackage[final]{colm2025_conference}

\usepackage{microtype}
\usepackage{hyperref}
\usepackage{url}
\usepackage{booktabs}
\usepackage{graphicx}
\usepackage{bookmark}
\usepackage{tcolorbox}
\usepackage{tabularx}
\usepackage{multirow, multicol}
\usepackage{enumitem}
\usepackage{amssymb}
\usepackage{xcolor}
\usepackage[T1]{fontenc}
\usepackage{lineno}

\definecolor{darkblue}{rgb}{0, 0, 0.5}
\hypersetup{colorlinks=true, citecolor=darkblue, linkcolor=darkblue, urlcolor=darkblue}

\newcolumntype{Y}{>{\centering\arraybackslash}X}

\newcommand{\rom}[1]{\lowercase\expandafter{\romannumeral #1\relax}}

\newcommand\red[1]{\textcolor{red}{#1}}

\newcommand{\n}{\textsc{\textsf{IRONIC}}}

\title{{\n}: Coherence-Aware Reasoning Chains for \\ Multi-Modal Sarcasm Detection}

\author{Aashish Anantha Ramakrishnan \\
College of Information Sciences and Technology \\
The Pennsylvania State University\\
\texttt{aza6352@psu.edu} \\
\AND
Aadarsh Anantha Ramakrishnan \\
Department of Computer Science and Engineering \\
National Institute of Technology, Tiruchirappalli\\
\texttt{aadarsh.ram@gmail.com} \\
\AND
Dongwon Lee \\
College of Information Sciences and Technology \\
The Pennsylvania State University\\
\texttt{dul13@psu.edu} \\
}

\begin{document}

\ifcolmsubmission
\linenumbers
\fi

\maketitle

\begin{abstract}
Interpreting figurative language such as sarcasm across multi-modal inputs presents unique challenges, often requiring task-specific fine-tuning and extensive reasoning steps. However, current Chain-of-Thought approaches do not efficiently leverage the same cognitive processes that enable humans to identify sarcasm. We present {\n}, an in-context learning framework that leverages Multi-modal Coherence Relations to analyze referential and pragmatic image-text linkages. Our experiments show that {\n} achieves state-of-the-art performance on zero-shot Multi-modal Sarcasm Detection across different baselines. This demonstrates the need for incorporating pragmatic insights into the design of multi-modal reasoning strategies.
\end{abstract}

\begin{figure*}[!ht]
    \centering
    \includegraphics[width=\linewidth]{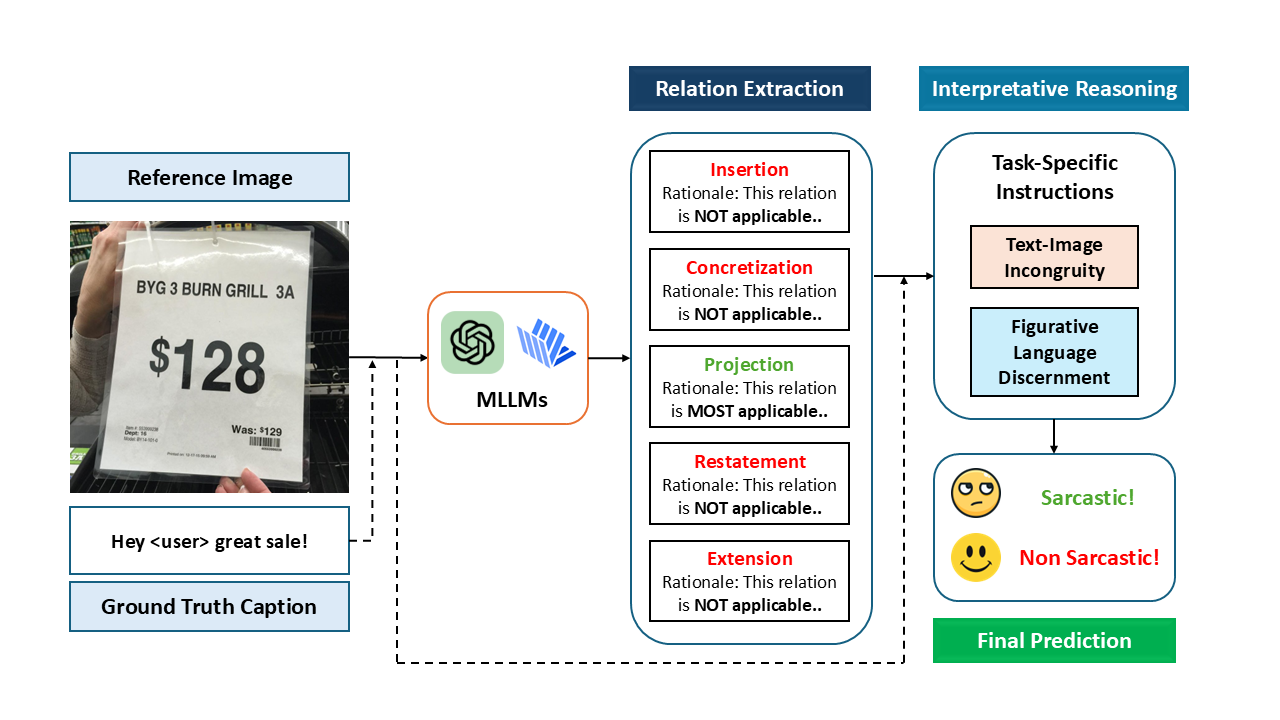}
    \caption{Overview of the proposed {\n} framework}
    \label{fig:framework}   
\end{figure*}

\section{Introduction}
\label{sec:intro}

Sarcasm can be defined as a form of verbal irony where the goal is to mock or convey contempt while the literal meaning of the words is different, often opposite to the intended meaning \citep{Camp2012-wc}. Multi-modal sarcasm utilizes the interplay between these visual and text cues to convey these intents.  The task of multi-modal sarcasm detection is helpful for various applications, including social media analysis, sentiment analysis, and human-computer interaction \citep{Pang2008-po}. With traditional supervised learning approaches demonstrating limited success in this domain, recent research has focused on leveraging Multi-modal Large Language Models (MLLMs) to improve performance \citep{Tang2024-mk}. These models have shown promise even in Zero-Shot settings, where they can generalize to new tasks without explicit training data. \textit{The complexity of multi-modal sarcasm detection stems from the fact that it incorporates a combination of semantic, pragmatic, and analogical cues, which are not always aligned with the logical and sequential nature of traditional reasoning strategies of language models} \citep{Yao2025-zr,Kumar2022-do}. This leads to models taking shortcuts such as over-reliance on uni-modal features and failing to discern sarcasm from other forms of figurative language \citep{Liang2022-mx,Qin2023-sy}.

Existing approaches for prompt-based reasoning in MLLMs have focused on leveraging different varieties of logical chains, such as Chain-of-Thought (CoT) prompting \citep{Wei2022-or}. Inspired by the success of text-only cognition-aware and linguistics-based reasoning strategies \citep{Wang2024-hx,Lee2025-ch}, we propose {\n} (expl\underline{I}cit \underline{R}easoning with c\underline{O}herence relatio\underline{N}s as cogn\underline{I}tive stru\underline{C}tures). {\n} leverages multi-modal Coherence Relations \citep{Anantha-Ramakrishnan2025-ce,Alikhani2020-nr} to provide a structured reasoning pathway for MLLMs to follow while identifying image-text incongruencies. Coherence Relations (CR) are a formal theory of discourse \citep{Hobbs1978-em} that provides a framework for understanding how different communicative components relate to each other \citep{Hobbs1978-em}. As a task-independent framework, CR-based reasoning can help generate informative rationales for multi-modal sarcasm detection. 

We analyze the capabilities of both proprietary (GPT-4o) and open-source models (InternVL3) in using Coherence Relations for reasoning \citep{Zhu2025-ua, OpenAI2024-hr}. Our findings on two different datasets: MMSD2.0 \citep{Qin2023-sy} and RedEval \citep{Tang2024-mk} show that the higher-order reasoning abilities of MLLMs play a crucial role in how they leverage CRs. {\n} enables more pragmatically-aware models such as GPT-4o \citep{OpenAI2024-hr} to achieve state-of-the-art performance on zero-shot multi-modal sarcasm detection tasks. The contributions of this work can be summarized as follows: 

\begin{itemize}[leftmargin=3.3mm]
  \item We present {\n}, a coherence-aware in-context reasoning framework that leverages Multi-modal Discourse Coherence Relations for effectively analyzing image-text incongruencies.
  \item {\n} enables MLLMs with emergent higher-order reasoning capabilities to outperform existing multi-step reasoning strategies such as CoT in zero-shot settings across different datasets.
  \item Our results demonstrate the importance of leveraging linguistic and cognitive insights into the design of multi-modal reasoning chains, improving the performance of MLLMs on pragmatic comprehension tasks.
\end{itemize}

\section{Methodology}
\label{sec:methodology}

The proposed {\n} framework, as shown in Figure \ref{fig:framework}, constructs a reasoning chain with 2 main components: (1) Relation Extraction (2) Interpretative Reasoning. We analyze related works on Sarcasm Detection and Image-Text Alignment in Appendix Section \ref{sec:related_work}.

\paragraph{Relation Extraction}
We utilize CRs that provide coverage of referential, analogical, and pragmatic reasoning types with varying levels of granularity. This includes cues that focus on both entity-level and scene-level features, supporting a comprehensive analysis of image-text relations. The selection of these CRs is supported by their applicability to analyzing image-text pairs from social media domains \citep{Xu2022-ie}, ensuring strong generalizability with the datasets utilized for multi-modal sarcasm detection. The outputs of this step include the predicted CRs and their supporting rationales for each sample. The predicted CR distributions and specific prompts are shown in the Appendix Sections \ref{sec:cr-dist} \& \ref{sec:prompt_templates}. Our framework incorporates 5 distinct CRs:

\begin{itemize}[leftmargin=3.3mm]
    \item \textbf{Insertion} (Entity-Level): The reader must infer the referent object or main entity from the image with no explicit mentions in the text.
    \item \textbf{Concretization} (Entity-Level): Identifies explicit referential alignment between image and text along with implied communicative intents.
    \item \textbf{Projection} (Entity-Level): Presents an analogical mapping between the main topic discussed in the text and the objects presented in the image. 
    \item \textbf{Restatement} (Scene-Level): Describes an explicit referential agreement with strong semantic overlap across modalities.
    \item \textbf{Extension} (Scene-Level): The text introduces narrative elements, speaker intent, or metaphorical framing that elaborates on the visual scene with information not directly inferable from the image alone.
    
\end{itemize} 

\paragraph{Interpretative Reasoning}
Although MLLMs show limited ability in explicit coherence understanding tasks such as predicting CRs \citep{Anantha-Ramakrishnan2025-ce, Jia2025-ho}, recent studies have shown that language models are effective in utilizing coherence-aware reasoning chains for downstream higher-order reasoning tasks \citep{Zhang2025-ns,Anantha-Ramakrishnan2025-je}. This motivates the second step of our reasoning chain, where we integrate the image-text pair, the predicted CR, and the generated rationale for distinguishing if a sample contains multi-modal sarcasm. In addition to these inputs, we provide task-specific instructions to discern simple humor or satirical commentary from sarcasm indicators such as mockery or irony concerning a specific referent. This enables our Interpretative Reasoning step to be robust, where the coherence-aware rationale is used as a supporting element while basing the final decision on the full multi-modal input. Thus, by providing both general coherence rationales and task-specific instructions, we develop a flexible prompting strategy that can handle inconsistencies in upstream reasoning steps.

\begin{table*}[!ht]
    \centering
    \small
        \begin{tabularx}{\linewidth}{@{} l | c | c | YYY | Y @{}}
        \toprule

        \textbf{Dataset} & \textbf{Model} & \textbf{Type} & \textbf{Accuracy} & \textbf{Precision} & \textbf{Recall} & \textbf{F1} \\
        \midrule

        \multirow{8}{*}{MMSD 2.0} & \multirow{4}{*}{GPT4o} & Zero-Shot & 0.7107 & 0.7952 & 0.7107 & 0.7024 \\
         & & Zero-Shot CoT & 0.7468 & \textbf{0.8096} & 0.7468 & 0.7430 \\
         & & $S^{3}$ CoT & 0.6559 & 0.7863 & 0.6559 & 0.6344 \\
         & & {\n} & \textbf{0.7659} & 0.7729 & \textbf{0.7659} & \textbf{0.7670} \\

         \cmidrule{2-7}

         & \multirow{4}{*}{InternVL3-14B} & Zero-Shot & 0.7393 & 0.7427 & 0.7393 & 0.7322 \\
         & & Zero-Shot CoT & \textbf{0.7406} & \textbf{0.7752} & \textbf{0.7406} & \textbf{0.7400} \\
         & & $S^{3}$ CoT & 0.6874 & 0.7727 & 0.6874 & 0.6775 \\
         & & {\n} & 0.7177 & 0.7294 & 0.7177 & 0.7032 \\

        \midrule
        \midrule

        \multirow{8}{*}{RedEval} & \multirow{4}{*}{GPT4o} & Zero-Shot & 0.7799 & 0.8423 & 0.7799 & 0.7807 \\
         & & Zero-Shot CoT & 0.7669 & 0.8290 & 0.7669 & 0.7677 \\
         & & $S^{3}$ CoT & 0.7002 & 0.8172 & 0.7002 & 0.6939 \\
         & & {\n} & \textbf{0.8406} & \textbf{0.8501} & \textbf{0.8406} & \textbf{0.8422} \\

         \cmidrule{2-7}

         & \multirow{4}{*}{InternVL3-14B} & Zero-Shot & \textbf{0.8327} & \textbf{0.8319} & \textbf{0.8327} & \textbf{0.8309} \\
         & & Zero-Shot CoT & 0.7420 & 0.8039 & 0.7420 & 0.7428 \\
         & & $S^{3}$ CoT & 0.7122 & 0.8201 & 0.7122 & 0.7075 \\
         & & {\n} & 0.8197 & 0.8188 & 0.8197 & 0.8175 \\

        \bottomrule

        \end{tabularx}
       \caption{Results for Multi-Modal Sarcasm Detection on MMSD2.0 and RedEval datasets}
        \label{table:metrics_tweets}
\end{table*}

\section{Experiments}
\label{sec:experiments}

\paragraph{Datasets}
We select two popular multi-modal sarcasm detection datasets: MMSD2.0 \citep{Qin2023-sy} and RedEval \citep{Tong2024-kx} for evaluating the effectiveness of {\n}. It builds on the popular Twitter-based MMSD dataset \citep{Cai2019-wa} by removing spurious cues such as hashtags and unreasonable labels through re-annotation.  MMSD2.0's test set contains 1037 and 1372 examples for the sarcasm and the non-sarcasm classes. RedEval was constructed for Out-of-Domain evaluation of models trained on Twitter-based datasets. Samples in RedEval are sourced from Reddit, with 395 sarcastic examples selected from the "sarcasm" subreddit and 609 negative examples from other subreddits.

\paragraph{Evaluation Metrics}
To measure the performance of MLLMs on multi-modal sarcasm detection, we use the F1 score (weighted) as our main metric since it is framed as a binary classification task. We also report overall Accuracy, Precision, and Recall for all models.

\paragraph{Baselines Compared} \label{sec:baselines}
To understand how MLLMs utilize different types of reasoning chains compared to {\n}, we include a variety of prompting strategies in our experiments. Zero-Shot serves as a simple baseline where only the input image and text are provided with a single task prompt for sarcasm prediction. Zero-Shot CoT requires the model to generate a step-by-step rationale before deciding whether sarcasm is present in the inputs. We also compare $S^{3}$ CoT, a strategy based on $S^{3}$ Agent \citep{Wang2024-vm} for multi-modal sarcasm detection. $S^{3}$ Agent analyzes image-text pairs from 3 critical perspectives: superficial expression, semantic information, and sentiment expression using a multi-agent framework. To ensure consistency, we combine the parallel agent prompts that generate 3 separate rationales into one single prompt, followed by a decision agent to predict the presence of sarcasm. This way, $S^{3}$ CoT is also a two-step strategy, serving as an alternate reasoning chain baseline.

\section{Results}
\label{sec:results} 

\paragraph{{\n} Improves over CoT Baselines}
From our presented results on both datasets, we observe that {\n} outperforms $S^{3}$ CoT across all settings and Zero-Shot CoT in 3 out of the 4 settings compared. With GPT-4o, we see an improvement of 3.23\% and 9.70\% on both MMSD2.0 and RedEval, respectively, compared to Zero-Shot CoT. Similarly, for InternVL3, we observe a 10.06\% improvement over Zero-Shot CoT in RedEval. {\n} when combined with GPT-4o beats all previous state-of-the-art Zero-Shot approaches for MLLM-based sarcasm detection on both datasets \citep{Zhang2025-ky, Tang2024-mk}. This demonstrates the benefits of a structured cognitive framework for sarcasm identification over empirical prompting strategies in Zero-Shot settings, supporting our initial hypothesis.

\paragraph{Challenges in Explicit Reasoning Incorporation}
An interesting observation that we find when testing InternVL3-14B, is the strong zero-shot performance without separate reasoning steps on both datasets. The introduction of reasoning chains, both general and coherence-aware, does not provide meaningful performance improvements. This suggests that the model may rely on implicit heuristics for sarcasm recognition. Another potential reason for this behavior may be the lack of capacity to generate high-quality intermediate rationales \cite{Anantha-Ramakrishnan2025-ce}. This indicates that larger models may demonstrate higher internal reasoning fidelity to make use of cognitive reasoning scaffolds such as CRs. 

\section{Conclusion}
\label{sec:conclusion}
We propose {\n}, a coherence-aware in-context prompting strategy inspired by the Theory of Discourse Coherence for multi-modal sarcasm detection. Our study covers both proprietary and open-source models across image-text pairs sourced from 2 different social media platforms. We observe that {\n} elicits improved reasoning capabilities from MLLMs compared to other CoT approaches. Our experiments also investigate the distinction between implicit pattern matching and explicit reasoning, demonstrating the importance of expanding model capabilities to include pragmatic reasoning, moving beyond factual and logical approaches. 

\section*{Limitations}
\label{sec:limitations}
Our current experimental setup for {\n} is limited to a couple of top-performing proprietary and open-source models. We are working on expanding our study to cover a more comprehensive set of models as a part of future work. Additionally, our evaluation strategy does not incorporate task-specific fine-tuning. Exploring multi-task fine-tuning strategies to improve both CR prediction and sarcasm recognition are promising research directions we aim to pursue. Finally, to facilitate a deeper analysis on the Relation Extraction step of {\n}, we plan to explore human evaluation approaches for rationale quality analysis, facilitating identification of reasoning gaps and hallucinations.

\section*{Ethics Statement}
With Multi-modal Sarcasm detection requiring implicit assumptions that leverage a model's world view, we acknowledge the potential negative impacts of unintended stereotypical biases learned by these models. Similarly, the types of CRs we analyze are optimized for image-text linkages present in English. This may result in unexpected behavior when evaluating sarcasm in multi-lingual or code-mixed settings. Thus, we advocate the safe use of these approaches with human-in-the-loop validation. Similarly, we release all code and prompt templates to support transparency, reproducibility, and further research on coherence-aware reasoning in multi-modal models.

\section*{Acknowledgments}
This research was in part supported by the U.S. National Science Foundation (NSF) award \#1820609. Part of the research results were obtained using the computational resources provided by CloudBank (\url{https://www.cloudbank.org/}), which was supported by the NSF award \#1925001.

\bibliography{colm2025_conference}

\begin{thebibliography}{37}
\providecommand{\natexlab}[1]{#1}
\providecommand{\url}[1]{\texttt{#1}}
\expandafter\ifx\csname urlstyle\endcsname\relax
  \providecommand{\doi}[1]{doi: #1}\else
  \providecommand{\doi}{doi: \begingroup \urlstyle{rm}\Url}\fi

\bibitem[Alikhani et~al.(2020)Alikhani, Sharma, Li, Soricut, and Stone]{Alikhani2020-nr}
Malihe Alikhani, Piyush Sharma, Shengjie Li, Radu Soricut, and Matthew Stone.
\newblock Cross-modal coherence modeling for caption generation.
\newblock In Dan Jurafsky, Joyce Chai, Natalie Schluter, and Joel Tetreault (eds.), \emph{Proceedings of the 58th Annual Meeting of the Association for Computational Linguistics}, pp.\  6525--6535, Stroudsburg, PA, USA, 2020. Association for Computational Linguistics.

\bibitem[Alikhani et~al.(2022)Alikhani, Han, Ravi, Kapadia, Pavlovic, and Stone]{Alikhani2022-ut}
Malihe Alikhani, Fangda Han, Hareesh Ravi, Mubbasir Kapadia, Vladimir Pavlovic, and Matthew Stone.
\newblock Cross-modal coherence for text-to-image retrieval.
\newblock \emph{Proc. Conf. AAAI Artif. Intell.}, 36\penalty0 (10):\penalty0 10427--10435, June 2022.

\bibitem[Anantha~Ramakrishnan et~al.(2024)Anantha~Ramakrishnan, Huang, and Lee]{Anantha-Ramakrishnan2024-sv}
Aashish Anantha~Ramakrishnan, Sharon~X Huang, and Dongwon Lee.
\newblock {ANNA}: Abstractive text-to-image synthesis with filtered news captions.
\newblock In \emph{The Third Workshop on Advances in Language and Vision Research}. Association for Computational Linguistics, 2024.

\bibitem[Anantha~Ramakrishnan et~al.(2025{\natexlab{a}})Anantha~Ramakrishnan, Anantha~Ramakrishnan, and Dongwon]{Anantha-Ramakrishnan2025-ce}
Aashish Anantha~Ramakrishnan, Aadarsh Anantha~Ramakrishnan, and Lee Dongwon.
\newblock {CORDIAL}: Can multimodal large language models effectively understand coherence relationships?
\newblock \emph{arXiv [cs.CL]}, February 2025{\natexlab{a}}.

\bibitem[Anantha~Ramakrishnan et~al.(2025{\natexlab{b}})Anantha~Ramakrishnan, Anantha~Ramakrishnan, and Lee]{Anantha-Ramakrishnan2025-je}
Aashish Anantha~Ramakrishnan, Aadarsh Anantha~Ramakrishnan, and Dongwon Lee.
\newblock {RONA}: Pragmatically diverse image captioning with coherence relations.
\newblock In \emph{Proceedings of the Fourth Workshop on Intelligent and Interactive Writing Assistants (In2Writing 2025)}, pp.\  74--86. Association for Computational Linguistics, May 2025{\natexlab{b}}.

\bibitem[Cai et~al.(2019)Cai, Cai, and Wan]{Cai2019-wa}
Yitao Cai, Huiyu Cai, and Xiaojun Wan.
\newblock Multi-modal sarcasm detection in twitter with hierarchical fusion model.
\newblock In \emph{Proceedings of the 57th Annual Meeting of the Association for Computational Linguistics}, pp.\  2506--2515, Stroudsburg, PA, USA, 2019. Association for Computational Linguistics.

\bibitem[Camp(2012)]{Camp2012-wc}
Elisabeth Camp.
\newblock Sarcasm, pretense, and the semantics/pragmatics distinction.
\newblock \emph{Nous}, 46\penalty0 (4):\penalty0 587--634, December 2012.

\bibitem[Castro et~al.(2019)Castro, Hazarika, Pérez-Rosas, Zimmermann, Mihalcea, and Poria]{Castro2019-xf}
Santiago Castro, Devamanyu Hazarika, Verónica Pérez-Rosas, Roger Zimmermann, Rada Mihalcea, and Soujanya Poria.
\newblock Towards multimodal sarcasm detection (an \_obviously\_ perfect paper).
\newblock In \emph{Proceedings of the 57th Annual Meeting of the Association for Computational Linguistics}, pp.\  4619--4629, Stroudsburg, PA, USA, 2019. Association for Computational Linguistics.

\bibitem[Das \& Clark(2018)Das and Clark]{Das2018-bs}
Dipto Das and Anthony~J Clark.
\newblock Sarcasm detection on facebook: a supervised learning approach.
\newblock In \emph{Proceedings of the 20th International Conference on Multimodal Interaction: Adjunct}, New York, NY, USA, October 2018. ACM.

\bibitem[Hobbs(1978)]{Hobbs1978-em}
Jerry~R Hobbs.
\newblock \emph{Why is discourse coherent?}, volume 176.
\newblock SRI International Menlo Park, CA, 1978.

\bibitem[Jia et~al.(2025)Jia, Zhang, Zhang, and Wan]{Jia2025-ho}
Boyu Jia, Junzhe Zhang, Huixuan Zhang, and Xiaojun Wan.
\newblock Exploring and evaluating multimodal knowledge reasoning consistency of multimodal large language models.
\newblock \emph{arXiv [cs.CL]}, March 2025.

\bibitem[Jia et~al.(2024)Jia, Xie, and Jing]{Jia2024-ym}
Mengzhao Jia, Can Xie, and Liqiang Jing.
\newblock Debiasing multimodal sarcasm detection with contrastive learning.
\newblock \emph{Proc. Conf. AAAI Artif. Intell.}, 38\penalty0 (16):\penalty0 18354--18362, March 2024.

\bibitem[Kumar et~al.(2022)Kumar, Kulkarni, Akhtar, and Chakraborty]{Kumar2022-do}
Shivani Kumar, Atharva Kulkarni, Md~Shad Akhtar, and Tanmoy Chakraborty.
\newblock When did you become so smart, oh wise one?! sarcasm explanation in multi-modal multi-party dialogues.
\newblock In \emph{Proceedings of the 60th Annual Meeting of the Association for Computational Linguistics (Volume 1: Long Papers)}, pp.\  5956--5968, Stroudsburg, PA, USA, 2022. Association for Computational Linguistics.

\bibitem[Lee et~al.(2025)Lee, Fong, Le, Shah, Han, and Zhu]{Lee2025-ch}
Joshua Lee, Wyatt Fong, Alexander Le, Sur Shah, Kevin Han, and Kevin Zhu.
\newblock Pragmatic metacognitive prompting improves {LLM} performance on sarcasm detection.
\newblock In \emph{Proceedings of the 1st Workshop on Computational Humor (CHum)}, pp.\  63--70, 2025.

\bibitem[Liang et~al.(2022)Liang, Lou, Li, Yang, Gui, He, Pei, and Xu]{Liang2022-mx}
Bin Liang, Chenwei Lou, Xiang Li, Min Yang, Lin Gui, Yulan He, Wenjie Pei, and Ruifeng Xu.
\newblock Multi-modal sarcasm detection via cross-modal graph convolutional network.
\newblock In \emph{Proceedings of the 60th Annual Meeting of the Association for Computational Linguistics (Volume 1: Long Papers)}, pp.\  1767--1777, Stroudsburg, PA, USA, 2022. Association for Computational Linguistics.

\bibitem[Liu et~al.(2021)Liu, Gong, Wu, Zhang, Su, and Liu]{Liu2021-hn}
Xingchao Liu, Chengyue Gong, Lemeng Wu, Shujian Zhang, Hao Su, and Qiang Liu.
\newblock {FuseDream}: Training-free text-to-image generation with improved {CLIP+GAN} space optimization.
\newblock \emph{arXiv:2112.01573 [cs]}, December 2021.

\bibitem[Mahajan \& Roth(2020)Mahajan and Roth]{Mahajan2020-sr}
Shweta Mahajan and Stefan Roth.
\newblock Diverse image captioning with context-object split latent spaces.
\newblock In \emph{Proceedings of the 34th International Conference on Neural Information Processing Systems}, NIPS '20, Red Hook, NY, USA, 2020. Curran Associates Inc.

\bibitem[{OpenAI} et~al.(2024){OpenAI}, Hurst, Lerer, Goucher, Perelman, Ramesh, Clark, Ostrow, Welihinda, Hayes, Radford, Mądry, Baker-Whitcomb, Beutel, Borzunov, Carney, Chow, Kirillov, Nichol, Paino, Renzin, Passos, Kirillov, Christakis, Conneau, Kamali, Jabri, Moyer, Tam, Crookes, Tootoochian, Tootoonchian, Kumar, Vallone, Karpathy, Braunstein, Cann, Codispoti, Galu, Kondrich, Tulloch, Mishchenko, Baek, Jiang, Pelisse, Woodford, Gosalia, Dhar, Pantuliano, Nayak, Oliver, Zoph, Ghorbani, Leimberger, Rossen, Sokolowsky, Wang, Zweig, Hoover, Samic, McGrew, Spero, Giertler, Cheng, Lightcap, Walkin, Quinn, Guarraci, Hsu, Kellogg, Eastman, Lugaresi, Wainwright, Bassin, Hudson, Chu, Nelson, Li, Shern, Conger, Barette, Voss, Ding, Lu, Zhang, Beaumont, Hallacy, Koch, Gibson, Kim, Choi, McLeavey, Hesse, Fischer, Winter, Czarnecki, Jarvis, Wei, Koumouzelis, Sherburn, Kappler, Levin, Levy, Carr, Farhi, Mely, Robinson, Sasaki, Jin, Valladares, Tsipras, Li, Nguyen, Findlay, Oiwoh, Wong, Asdar, Proehl, Yang, Antonow,
  Kramer, Peterson, Sigler, Wallace, Brevdo, Mays, Khorasani, Such, Raso, Zhang, von Lohmann, Sulit, Goh, Oden, Salmon, Starace, Brockman, Salman, Bao, Hu, Wong, Wang, Schmidt, Whitney, Jun, Kirchner, Pinto, Ren, Chang, Chung, Kivlichan, O'Connell, O'Connell, Osband, Silber, Sohl, Okuyucu, Lan, Kostrikov, Sutskever, Kanitscheider, Gulrajani, Coxon, Menick, Pachocki, Aung, Betker, Crooks, Lennon, Kiros, Leike, Park, Kwon, Phang, Teplitz, Wei, Wolfe, Chen, Harris, Varavva, Lee, Shieh, Lin, Yu, Weng, Tang, Yu, Jang, Candela, Beutler, Landers, Parish, Heidecke, Schulman, Lachman, McKay, Uesato, Ward, Kim, Huizinga, Sitkin, Kraaijeveld, Gross, Kaplan, Snyder, Achiam, Jiao, Lee, Zhuang, Harriman, Fricke, Hayashi, Singhal, Shi, Karthik, Wood, Rimbach, Hsu, Nguyen, Gu-Lemberg, Button, Liu, Howe, Muthukumar, Luther, Ahmad, Kai, Itow, Workman, Pathak, Chen, Jing, Guy, Fedus, Zhou, Mamitsuka, Weng, McCallum, Held, Ouyang, Feuvrier, Zhang, Kondraciuk, Kaiser, Hewitt, Metz, Doshi, Aflak, Simens, Boyd, Thompson, Dukhan,
  Chen, Gray, Hudnall, Zhang, Aljubeh, Litwin, Zeng, Johnson, Shetty, Gupta, Shah, Yatbaz, Yang, Zhong, Glaese, Chen, Janner, Lampe, Petrov, Wu, Wang, Fradin, Pokrass, Castro, de~Castro, Pavlov, Brundage, Wang, Khan, Murati, Bavarian, Lin, Yesildal, Soto, Gimelshein, Cone, Staudacher, Summers, LaFontaine, Chowdhury, Ryder, Stathas, Turley, Tezak, Felix, Kudige, Keskar, Deutsch, Bundick, Puckett, Nachum, Okelola, Boiko, Murk, Jaffe, Watkins, Godement, Campbell-Moore, Chao, McMillan, Belov, Su, Bak, Bakkum, Deng, Dolan, Hoeschele, Welinder, Tillet, Pronin, Tillet, Dhariwal, Yuan, Dias, Lim, Arora, Troll, Lin, Lopes, Puri, Miyara, Leike, Gaubert, Zamani, Wang, Donnelly, Honsby, Smith, Sahai, Ramchandani, Huet, Carmichael, Zellers, Chen, Chen, Nigmatullin, Cheu, Jain, Altman, Schoenholz, Toizer, Miserendino, Agarwal, Culver, Ethersmith, Gray, Grove, Metzger, Hermani, Jain, Zhao, Wu, Jomoto, Wu, {Shuaiqi}, {Xia}, Phene, Papay, Narayanan, Coffey, Lee, Hall, Balaji, Broda, Stramer, Xu, Gogineni, Christianson,
  Sanders, Patwardhan, Cunninghman, Degry, Dimson, Raoux, Shadwell, Zheng, Underwood, Markov, Sherbakov, Rubin, Stasi, Kaftan, Heywood, Peterson, Walters, Eloundou, Qi, Moeller, Monaco, Kuo, Fomenko, Chang, Zheng, Zhou, Manassra, Sheu, Zaremba, Patil, Qian, Kim, Cheng, Zhang, He, Zhang, Jin, Dai, and Malkov]{OpenAI2024-hr}
{OpenAI}, Aaron Hurst, Adam Lerer, Adam~P Goucher, Adam Perelman, Aditya Ramesh, Aidan Clark, A~J Ostrow, Akila Welihinda, Alan Hayes, Alec Radford, Aleksander Mądry, Alex Baker-Whitcomb, Alex Beutel, Alex Borzunov, Alex Carney, Alex Chow, Alex Kirillov, Alex Nichol, Alex Paino, Alex Renzin, Alex~Tachard Passos, Alexander Kirillov, Alexi Christakis, Alexis Conneau, Ali Kamali, Allan Jabri, Allison Moyer, Allison Tam, Amadou Crookes, Amin Tootoochian, Amin Tootoonchian, Ananya Kumar, Andrea Vallone, Andrej Karpathy, Andrew Braunstein, Andrew Cann, Andrew Codispoti, Andrew Galu, Andrew Kondrich, Andrew Tulloch, Andrey Mishchenko, Angela Baek, Angela Jiang, Antoine Pelisse, Antonia Woodford, Anuj Gosalia, Arka Dhar, Ashley Pantuliano, Avi Nayak, Avital Oliver, Barret Zoph, Behrooz Ghorbani, Ben Leimberger, Ben Rossen, Ben Sokolowsky, Ben Wang, Benjamin Zweig, Beth Hoover, Blake Samic, Bob McGrew, Bobby Spero, Bogo Giertler, Bowen Cheng, Brad Lightcap, Brandon Walkin, Brendan Quinn, Brian Guarraci, Brian Hsu,
  Bright Kellogg, Brydon Eastman, Camillo Lugaresi, Carroll Wainwright, Cary Bassin, Cary Hudson, Casey Chu, Chad Nelson, Chak Li, Chan~Jun Shern, Channing Conger, Charlotte Barette, Chelsea Voss, Chen Ding, Cheng Lu, Chong Zhang, Chris Beaumont, Chris Hallacy, Chris Koch, Christian Gibson, Christina Kim, Christine Choi, Christine McLeavey, Christopher Hesse, Claudia Fischer, Clemens Winter, Coley Czarnecki, Colin Jarvis, Colin Wei, Constantin Koumouzelis, Dane Sherburn, Daniel Kappler, Daniel Levin, Daniel Levy, David Carr, David Farhi, David Mely, David Robinson, David Sasaki, Denny Jin, Dev Valladares, Dimitris Tsipras, Doug Li, Duc~Phong Nguyen, Duncan Findlay, Edede Oiwoh, Edmund Wong, Ehsan Asdar, Elizabeth Proehl, Elizabeth Yang, Eric Antonow, Eric Kramer, Eric Peterson, Eric Sigler, Eric Wallace, Eugene Brevdo, Evan Mays, Farzad Khorasani, Felipe~Petroski Such, Filippo Raso, Francis Zhang, Fred von Lohmann, Freddie Sulit, Gabriel Goh, Gene Oden, Geoff Salmon, Giulio Starace, Greg Brockman, Hadi
  Salman, Haiming Bao, Haitang Hu, Hannah Wong, Haoyu Wang, Heather Schmidt, Heather Whitney, Heewoo Jun, Hendrik Kirchner, Henrique Ponde de~Oliveira Pinto, Hongyu Ren, Huiwen Chang, Hyung~Won Chung, Ian Kivlichan, Ian O'Connell, Ian O'Connell, Ian Osband, Ian Silber, Ian Sohl, Ibrahim Okuyucu, Ikai Lan, Ilya Kostrikov, Ilya Sutskever, Ingmar Kanitscheider, Ishaan Gulrajani, Jacob Coxon, Jacob Menick, Jakub Pachocki, James Aung, James Betker, James Crooks, James Lennon, Jamie Kiros, Jan Leike, Jane Park, Jason Kwon, Jason Phang, Jason Teplitz, Jason Wei, Jason Wolfe, Jay Chen, Jeff Harris, Jenia Varavva, Jessica~Gan Lee, Jessica Shieh, Ji~Lin, Jiahui Yu, Jiayi Weng, Jie Tang, Jieqi Yu, Joanne Jang, Joaquin~Quinonero Candela, Joe Beutler, Joe Landers, Joel Parish, Johannes Heidecke, John Schulman, Jonathan Lachman, Jonathan McKay, Jonathan Uesato, Jonathan Ward, Jong~Wook Kim, Joost Huizinga, Jordan Sitkin, Jos Kraaijeveld, Josh Gross, Josh Kaplan, Josh Snyder, Joshua Achiam, Joy Jiao, Joyce Lee, Juntang
  Zhuang, Justyn Harriman, Kai Fricke, Kai Hayashi, Karan Singhal, Katy Shi, Kavin Karthik, Kayla Wood, Kendra Rimbach, Kenny Hsu, Kenny Nguyen, Keren Gu-Lemberg, Kevin Button, Kevin Liu, Kiel Howe, Krithika Muthukumar, Kyle Luther, Lama Ahmad, Larry Kai, Lauren Itow, Lauren Workman, Leher Pathak, Leo Chen, Li~Jing, Lia Guy, Liam Fedus, Liang Zhou, Lien Mamitsuka, Lilian Weng, Lindsay McCallum, Lindsey Held, Long Ouyang, Louis Feuvrier, Lu~Zhang, Lukas Kondraciuk, Lukasz Kaiser, Luke Hewitt, Luke Metz, Lyric Doshi, Mada Aflak, Maddie Simens, Madelaine Boyd, Madeleine Thompson, Marat Dukhan, Mark Chen, Mark Gray, Mark Hudnall, Marvin Zhang, Marwan Aljubeh, Mateusz Litwin, Matthew Zeng, Max Johnson, Maya Shetty, Mayank Gupta, Meghan Shah, Mehmet Yatbaz, Meng~Jia Yang, Mengchao Zhong, Mia Glaese, Mianna Chen, Michael Janner, Michael Lampe, Michael Petrov, Michael Wu, Michele Wang, Michelle Fradin, Michelle Pokrass, Miguel Castro, Miguel Oom~Temudo de~Castro, Mikhail Pavlov, Miles Brundage, Miles Wang, Minal
  Khan, Mira Murati, Mo~Bavarian, Molly Lin, Murat Yesildal, Nacho Soto, Natalia Gimelshein, Natalie Cone, Natalie Staudacher, Natalie Summers, Natan LaFontaine, Neil Chowdhury, Nick Ryder, Nick Stathas, Nick Turley, Nik Tezak, Niko Felix, Nithanth Kudige, Nitish Keskar, Noah Deutsch, Noel Bundick, Nora Puckett, Ofir Nachum, Ola Okelola, Oleg Boiko, Oleg Murk, Oliver Jaffe, Olivia Watkins, Olivier Godement, Owen Campbell-Moore, Patrick Chao, Paul McMillan, Pavel Belov, Peng Su, Peter Bak, Peter Bakkum, Peter Deng, Peter Dolan, Peter Hoeschele, Peter Welinder, Phil Tillet, Philip Pronin, Philippe Tillet, Prafulla Dhariwal, Qiming Yuan, Rachel Dias, Rachel Lim, Rahul Arora, Rajan Troll, Randall Lin, Rapha~Gontijo Lopes, Raul Puri, Reah Miyara, Reimar Leike, Renaud Gaubert, Reza Zamani, Ricky Wang, Rob Donnelly, Rob Honsby, Rocky Smith, Rohan Sahai, Rohit Ramchandani, Romain Huet, Rory Carmichael, Rowan Zellers, Roy Chen, Ruby Chen, Ruslan Nigmatullin, Ryan Cheu, Saachi Jain, Sam Altman, Sam Schoenholz, Sam
  Toizer, Samuel Miserendino, Sandhini Agarwal, Sara Culver, Scott Ethersmith, Scott Gray, Sean Grove, Sean Metzger, Shamez Hermani, Shantanu Jain, Shengjia Zhao, Sherwin Wu, Shino Jomoto, Shirong Wu, {Shuaiqi}, {Xia}, Sonia Phene, Spencer Papay, Srinivas Narayanan, Steve Coffey, Steve Lee, Stewart Hall, Suchir Balaji, Tal Broda, Tal Stramer, Tao Xu, Tarun Gogineni, Taya Christianson, Ted Sanders, Tejal Patwardhan, Thomas Cunninghman, Thomas Degry, Thomas Dimson, Thomas Raoux, Thomas Shadwell, Tianhao Zheng, Todd Underwood, Todor Markov, Toki Sherbakov, Tom Rubin, Tom Stasi, Tomer Kaftan, Tristan Heywood, Troy Peterson, Tyce Walters, Tyna Eloundou, Valerie Qi, Veit Moeller, Vinnie Monaco, Vishal Kuo, Vlad Fomenko, Wayne Chang, Weiyi Zheng, Wenda Zhou, Wesam Manassra, Will Sheu, Wojciech Zaremba, Yash Patil, Yilei Qian, Yongjik Kim, Youlong Cheng, Yu~Zhang, Yuchen He, Yuchen Zhang, Yujia Jin, Yunxing Dai, and Yury Malkov.
\newblock {GPT}-{4o} system card.
\newblock \emph{arXiv [cs.CL]}, October 2024.

\bibitem[Pan et~al.(2020)Pan, Lin, Fu, Qi, and Wang]{Pan2020-uz}
Hongliang Pan, Zheng Lin, Peng Fu, Yatao Qi, and Weiping Wang.
\newblock Modeling intra and inter-modality incongruity for multi-modal sarcasm detection.
\newblock In \emph{Findings of the Association for Computational Linguistics: EMNLP 2020}, pp.\  1383--1392, Stroudsburg, PA, USA, November 2020. Association for Computational Linguistics.

\bibitem[Pang \& Lee(2008)Pang and Lee]{Pang2008-po}
Bo~Pang and Lillian Lee.
\newblock Opinion mining and sentiment analysis.
\newblock \emph{Found. Trends® Inf. Retr.}, 2\penalty0 (1-2):\penalty0 1--135, July 2008.

\bibitem[Qin et~al.(2023)Qin, Huang, Chen, Cai, Zhang, Liang, Che, and Xu]{Qin2023-sy}
Libo Qin, Shijue Huang, Qiguang Chen, Chenran Cai, Yudi Zhang, Bin Liang, Wanxiang Che, and Ruifeng Xu.
\newblock {MMSD2}.0: Towards a reliable multi-modal sarcasm detection system.
\newblock In \emph{Findings of the Association for Computational Linguistics: ACL 2023}, pp.\  10834--10845, Stroudsburg, PA, USA, 2023. Association for Computational Linguistics.

\bibitem[Radford et~al.(2021)Radford, Kim, Hallacy, Ramesh, Goh, Agarwal, Sastry, Askell, Mishkin, Clark, Krueger, and Sutskever]{Radford2021-ro}
Alec Radford, Jong~Wook Kim, Chris Hallacy, Aditya Ramesh, Gabriel Goh, Sandhini Agarwal, Girish Sastry, Amanda Askell, Pamela Mishkin, Jack Clark, Gretchen Krueger, and Ilya Sutskever.
\newblock Learning transferable visual models from natural language supervision.
\newblock \emph{arXiv:2103.00020 [cs]}, February 2021.

\bibitem[Ray et~al.(2022)Ray, Mishra, Nunna, and Bhattacharyya]{Ray2022-mi}
Anupama Ray, Shubham Mishra, Apoorva Nunna, and Pushpak Bhattacharyya.
\newblock A multimodal corpus for emotion recognition in sarcasm.
\newblock In \emph{Proceedings of the Thirteenth Language Resources and Evaluation Conference}, pp.\  6992--7003, 2022.

\bibitem[Schifanella et~al.(2016)Schifanella, de~Juan, Tetreault, and Cao]{Schifanella2016-ec}
Rossano Schifanella, Paloma de~Juan, Joel Tetreault, and Liangliang Cao.
\newblock Detecting sarcasm in multimodal social platforms.
\newblock In \emph{Proceedings of the 24th ACM international conference on Multimedia}, New York, NY, USA, October 2016. ACM.

\bibitem[Sosea et~al.(2021)Sosea, Sirbu, Caragea, Caragea, and Rebedea]{Sosea2021-hr}
Tiberiu Sosea, Iustin Sirbu, Cornelia Caragea, Doina Caragea, and Traian Rebedea.
\newblock Using the image-text relationship to improve multimodal disaster tweet classification.
\newblock \emph{Int Conf Inf Syst Crisis Response Manag}, pp.\  691--704, 2021.

\bibitem[Tang et~al.(2024)Tang, Lin, Yan, and Li]{Tang2024-mk}
Binghao Tang, Boda Lin, Haolong Yan, and Si~Li.
\newblock Leveraging generative large language models with visual instruction and demonstration retrieval for multimodal sarcasm detection.
\newblock In \emph{Proceedings of the 2024 Conference of the North American Chapter of the Association for Computational Linguistics: Human Language Technologies (Volume 1: Long Papers)}, pp.\  1732--1742, Stroudsburg, PA, USA, 2024. Association for Computational Linguistics.

\bibitem[Tong et~al.(2024)Tong, Liu, Zhai, Ma, LeCun, and Xie]{Tong2024-kx}
Shengbang Tong, Zhuang Liu, Yuexiang Zhai, Yi~Ma, Yann LeCun, and Saining Xie.
\newblock Eyes wide shut? exploring the visual shortcomings of {MultiModal} {LLMs}.
\newblock \emph{Proc. IEEE Comput. Soc. Conf. Comput. Vis. Pattern Recognit.}, pp.\  9568--9578, January 2024.

\bibitem[Vempala \& Preoţiuc-Pietro(2019)Vempala and Preoţiuc-Pietro]{Vempala2019-lh}
Alakananda Vempala and Daniel Preoţiuc-Pietro.
\newblock Categorizing and inferring the relationship between the text and image of twitter posts.
\newblock In \emph{Proceedings of the 57th Annual Meeting of the Association for Computational Linguistics}, pp.\  2830--2840, Stroudsburg, PA, USA, July 2019. Association for Computational Linguistics.

\bibitem[Wang et~al.(2024)Wang, Zhang, Fei, Chen, Wang, Si, Lu, Li, and Qin]{Wang2024-vm}
Peng Wang, Yongheng Zhang, Hao Fei, Qiguang Chen, Yukai Wang, Jiasheng Si, Wenpeng Lu, Min Li, and Libo Qin.
\newblock {S} $^{3}$ agent: Unlocking the power of {VLLM} for zero-shot multi-modal sarcasm detection.
\newblock \emph{ACM Trans. Multimed. Comput. Commun. Appl.}, August 2024.

\bibitem[Wang \& Zhao(2024)Wang and Zhao]{Wang2024-hx}
Yuqing Wang and Yun Zhao.
\newblock Metacognitive prompting improves understanding in large language models.
\newblock In \emph{Proceedings of the 2024 Conference of the North American Chapter of the Association for Computational Linguistics: Human Language Technologies (Volume 1: Long Papers)}, pp.\  1914--1926, Stroudsburg, PA, USA, 2024. Association for Computational Linguistics.

\bibitem[Wei et~al.(2022)Wei, Wang, Schuurmans, Bosma, Xia, Chi, Le, Zhou, and {Others}]{Wei2022-or}
Jason Wei, Xuezhi Wang, Dale Schuurmans, Maarten Bosma, Fei Xia, Ed~Chi, Quoc~V Le, Denny Zhou, and {Others}.
\newblock Chain-of-thought prompting elicits reasoning in large language models.
\newblock \emph{Advances in neural information processing systems}, 35:\penalty0 24824--24837, 2022.

\bibitem[Wu et~al.(2021)Wu, Zhao, Lu, Qin, Wu, Sheng, and Li]{Wu2021-ob}
Yang Wu, Yanyan Zhao, Xin Lu, Bing Qin, Yin Wu, Jian Sheng, and Jinlong Li.
\newblock Modeling incongruity between modalities for multimodal sarcasm detection.
\newblock \emph{IEEE Multimed.}, 28\penalty0 (2):\penalty0 86--95, April 2021.

\bibitem[Xu et~al.(2022)Xu, Tan, Li, and Li]{Xu2022-ie}
Chunpu Xu, Hanzhuo Tan, Jing Li, and Piji Li.
\newblock Understanding social media cross-modality discourse in linguistic space.
\newblock In Yoav Goldberg, Zornitsa Kozareva, and Yue Zhang (eds.), \emph{Findings of the Association for Computational Linguistics: EMNLP 2022}, pp.\  2459--2471, Stroudsburg, PA, USA, 2022. Association for Computational Linguistics.

\bibitem[Yao et~al.(2025)Yao, Zhang, Li, and Qin]{Yao2025-zr}
Ben Yao, Yazhou Zhang, Qiuchi Li, and Jing Qin.
\newblock Is sarcasm detection a step-by-step reasoning process in large language models?
\newblock \emph{Proc. Conf. AAAI Artif. Intell.}, 39\penalty0 (24):\penalty0 25651--25659, April 2025.

\bibitem[Zhang et~al.(2025{\natexlab{a}})Zhang, Wang, Ren, Jiang, Wang, and Liu]{Zhang2025-ns}
Jinghan Zhang, Xiting Wang, Weijieying Ren, Lu~Jiang, Dongjie Wang, and Kunpeng Liu.
\newblock {RATT}: A thought structure for coherent and correct {LLM} reasoning.
\newblock \emph{Proc. Conf. AAAI Artif. Intell.}, 39\penalty0 (25):\penalty0 26733--26741, April 2025{\natexlab{a}}.

\bibitem[Zhang et~al.(2025{\natexlab{b}})Zhang, Zou, Wang, and Qin]{Zhang2025-ky}
Yazhou Zhang, Chunwang Zou, Bo~Wang, and Jing Qin.
\newblock Commander-{GPT}: Fully unleashing the sarcasm detection capability of multi-modal large language models.
\newblock \emph{arXiv [cs.CL]}, March 2025{\natexlab{b}}.

\bibitem[Zhu et~al.(2025)Zhu, Wang, Chen, Liu, Ye, Gu, Duan, Tian, Su, Shao, Gao, Cui, Cao, Liu, Wei, Zhang, Wang, Xu, Li, Wang, Chen, Li, He, Jiang, Luo, Wang, He, Shi, Zhang, Shao, He, Xiong, Qu, Sun, Jiao, Lv, Wu, Zhang, Deng, Ge, Chen, Wang, Dou, Lu, Zhu, Lu, Lin, Qiao, Dai, and Wang]{Zhu2025-ua}
Jinguo Zhu, Weiyun Wang, Zhe Chen, Zhaoyang Liu, Shenglong Ye, Lixin Gu, Yuchen Duan, Hao Tian, Weijie Su, Jie Shao, Zhangwei Gao, Erfei Cui, Yue Cao, Yangzhou Liu, Xingguang Wei, Hongjie Zhang, Haomin Wang, Weiye Xu, Hao Li, Jiahao Wang, Dengnian Chen, Songze Li, Yinan He, Tan Jiang, Jiapeng Luo, Yi~Wang, Conghui He, Botian Shi, Xingcheng Zhang, Wenqi Shao, Junjun He, Yingtong Xiong, Wenwen Qu, Peng Sun, Penglong Jiao, Han Lv, Lijun Wu, Kaipeng Zhang, Huipeng Deng, Jiaye Ge, Kai Chen, Limin Wang, Min Dou, Lewei Lu, Xizhou Zhu, Tong Lu, Dahua Lin, Yu~Qiao, Jifeng Dai, and Wenhai Wang.
\newblock {InternVL3}: Exploring advanced training and test-time recipes for open-source multimodal models.
\newblock \emph{arXiv [cs.CV]}, April 2025.

\end{thebibliography}
\bibliographystyle{colm2025_conference}

\newpage
\appendix
\section*{Appendix}
\label{sec:appendix}

\section{Related Work}
\label{sec:related_work}

\subsection{Multi-Modal Sarcasm Detection}
Initial approaches for multi-modal sarcasm detection utilized supervised deep learning models to incorporate image and text features \citep{Schifanella2016-ec, Das2018-bs}. Attention mechanisms further improved their abilities to associate image-text features across modalities \citep{Pan2020-uz, Wu2021-ob}. This led to the incorporation of more modalities such as audio, video, and facial expressions \citep{Castro2019-xf, Ray2022-mi}. However, models still struggled to comprehend complex relationships between image-text modalities required for sarcasm understanding. With the popularity of pre-training architectures such as CLIP \citep{Radford2021-ro} and MLLMs, the capabilities of these models in identifying sarcastic content expanded further \citep{Jia2024-ym}. Prompting-based approaches gained prominence due to their ability to generalize across domains without task-specific fine-tuning \citep{Wang2024-vm, Zhang2025-ky}.

\subsection{Image-Text Relation Understanding}
Traditional multi-modal models worked solely on the assumption that information across modalities is highly correlated with each other \citep{Radford2021-ro}, with tasks such as image captioning and image generation leveraging their strong semantic overlap \citep{Mahajan2020-sr, Liu2021-hn}. However, as these multi-modal models expanded their applications across different task domains such as news \citep{Anantha-Ramakrishnan2024-sv} and social media \citep{Xu2022-ie, Vempala2019-lh}, the need for enhanced reasoning beyond semantic overlap became clear. Coherence Relations offer a cohesive framework for connecting different types of reasoning across multiple modalities and have improved model comprehension capabilities across different domains \citep{Alikhani2020-nr, Alikhani2022-ut, Xu2022-ie, Sosea2021-hr}. Motivated by these insights, we analyze the incorporation of CRs as a cognitive reasoning scaffold for multi-modal sarcasm detection.

\section{Distribution of Predicted Coherence Relations}
\label{sec:cr-dist}

As an extended analysis, we present the distribution of predicted Coherence Relations by GPT4o coupled with {\n} in Tables \ref{table:mmsd-dist} and \ref{table:redeval-dist}. This distribution has been constructed by parsing the Relation Extraction reasoning step. In MMSD 2.0 and RedEval, eight and two image-text pairs respectively do not have predicted Coherence Relations because their rationales exceeded the maximum token limit.

\begin{table}[h!]
\centering
\begin{tabular}{lcc}
\hline
\textbf{CR} & \textbf{Non-Sarcastic} & \textbf{Sarcastic} \\
\hline
Insertion        & 122 & 75  \\
Extension        & 248 & 297 \\
Concretization   & 84  & 39  \\
Projection       & 458 & 283 \\
Restatement      & 456 & 339 \\
\hline
\end{tabular}
\caption{Distribution of CRs in the MMSD 2.0 Dataset}
\label{table:mmsd-dist}
\end{table}

\begin{table}[h!]
\centering
\begin{tabular}{lcc}
\hline
\textbf{CR} & \textbf{Non-Sarcastic} & \textbf{Sarcastic} \\
\hline
Insertion        & 51  & 24  \\
Extension        & 96  & 129 \\
Concretization   & 27  & 16  \\
Projection       & 101 & 75  \\
Restatement      & 334 & 149 \\
\hline
\end{tabular}
\caption{Distribution of CRs in the RedEval Dataset}
\label{table:redeval-dist}
\end{table}

\section{Model Availability}

This section focuses on the details regarding model availability and parameters used for evaluation. For all models, we set temperature to $0$ or \texttt{do\_sample=False}, maximum output tokens to $512$ and the random seed set to $42$, wherever possible to ensure reproducibility.

\subsection{Proprietary Models}
\paragraph{OpenAI GPT:} We access the GPT-4o model via a custom deployment using Azure OpenAI. We evaluate \texttt{gpt-4o-2024-11-20} with a custom safety filter to restrict content of \textit{high} severity level. We also use the public version of GPT-4o, via the official OpenAI API.

\subsection{Open-Source Models}
We evaluate InternVL3 14B MLLM (\texttt{OpenGVLab/InternVL3-14B} on Huggingface) using the VLLM \footnote{\url{https://github.com/vllm-project/vllm}} framework.

\section{Prompt Templates}
\label{sec:prompt_templates}

We use various prompt templates with different system/user messages for {\n} and the baselines mentioned in Section \ref{sec:baselines}. The different prompts and system messages used are present in the appendix.

\section{Postprocessing MLLM Responses}
Since both datasets feature image-caption pairs on a wide selection of topics, around \textbf{50+ images} were flagged by Azure GPT4o's strict safety filter or rejected by the model during evaluation. To ensure test set consistency, we decided to rerun these samples using the official OpenAI API instead and were able to get results.

\begin{figure}[ht]
    \centering
    \begin{tcolorbox}[title={System Message for Baseline Zero-Shot/CoT Evaluation}, after skip=0pt, boxsep=5pt, width=\linewidth]

    You are an expert linguist. Your task is to analyze a image-text pair and determine if it is 'sarcastic' or 'non-sarcastic'.
    
    \end{tcolorbox}
\end{figure}

\begin{figure}[ht]
    \centering
    \begin{tcolorbox}[title={System Message for {\n}}, colframe = blue!30, colback = blue!10, coltitle = blue!20!black, after skip=0pt, boxsep=5pt, width=\textwidth]

    You are an expert linguist, and your task is to analyze and predict the most applicable Coherence Relations for image-text pairs. A coherence relation describes the structural, logical, and purposeful relationships between an image and its caption, capturing the author's intent. \\
    
    These are the possible coherence relations you can assign to an image-text pair:
    
    - Insertion: The salient object described in the image is not explicitly mentioned in the text. \\
    - Concretization: Both the text and image contain a mention of the main visual entity. \\
    - Projection: The main entity mentioned in the text is implicitly related to the visual objects present in the image. \\
    - Restatement: The text directly describes the image contents. \\
    - Extension: The image expands upon the story or idea in the text, presenting new elements or elaborations, effectively filling in narrative gaps left by the text.
    
    \end{tcolorbox}
\end{figure}

\begin{figure}[ht]
    \centering
    \begin{tcolorbox}[title={Baseline Zero-Shot Prompt}, colframe = red!30, colback = red!10, coltitle = red!20!black, after skip=0pt, boxsep=5pt, width=\textwidth]
    
    \textbf{System} \\
    <insert-system-message> \\
    
    \textbf{User} \\
    Output 0 if it does not contain sarcastic content, and 1 if it does. Do not add anything else in your response. \\

    \textbf{<insert-image-text-pair>} \\

    \end{tcolorbox}
\end{figure}

\begin{figure}[ht]
    \centering
    \begin{tcolorbox}[title={Baseline Zero-Shot CoT Prompt}, colframe = red!30, colback = red!10, coltitle = red!20!black, after skip=0pt, boxsep=5pt, width=\textwidth]
    
    \textbf{System} \\
    <insert-system-message> \\
    
    \textbf{User} \\
    Let's think \red{step-by-step} and analyze the relationship between the text and image carefully. \\

    \textbf{<insert-image-text-pair>} \\

    \textbf{Assistant} \\
    Analysis: <add-analysis-from-model> \\
    
    \textbf{User} \\
    Using your rationale, please determine if the image-text pair is sarcastic in nature. Output 0 if it does not contain sarcastic content, and 1 if it does. Do not add anything else in your response.

    \end{tcolorbox}
\end{figure}

\begin{figure}[ht]
    \centering
    \begin{tcolorbox}[title={{\n} Prompt}, colframe = green!30, colback = green!10, coltitle = green!10!black, after skip=0pt, boxsep=5pt, width=\textwidth]

    \textbf{System} \\
    <insert-system-message> \\
    
    \textbf{User} \\
    Let's think \red{step-by-step} and analyze the relationship between the text and image carefully. Please also identify the most appropriate coherence relation between the text and image. \\

    \textbf{<insert-image-text-pair>} \\

    \textbf{Assistant} \\
    Analysis: <add-analysis-from-model> \\

    \textbf{User} \\
    Using the coherence relation analysis and the input image-text pair, please determine if the content contains multi-modal sarcasm. Content where both image and text together create a sarcastic effect is considered multi-modal sarcasm, do not focus on the sarcasm in only the image or text. Clearly distinguish simple humor and satire which is not sarcastic from mockery and irony which is sarcastic. \\

    Output 0 if it does not contain multi-modal sarcasm, and 1 if it does. Do not add anything else in your response.

    \end{tcolorbox}
\end{figure}

\begin{figure}[ht]
    \centering
    \begin{tcolorbox}[title={$S^{3}$ Agent Prompt}, colframe = orange!30, colback = orange!10, coltitle = orange!20!black, after skip=0pt, boxsep=5pt, width=\textwidth]

    \textbf{User} \\
    Given the following image and text, please judge whether there is sarcasm based on the 3 perspectives below.

    \begin{enumerate}
        \item \textbf{Superficial Expression:} This includes detect underlying critiques in contexts through image-text discrepancies.
        \item \textbf{Semantic Information:} This includes detect extreme portrayals and metaphorical in contexts through image-text semantic.
        \item \textbf{Sentiment Expression:} This includes detect criticize emotion on specific subjects or behaviors in the content.
    \end{enumerate}

    Without considering conclusions drawn solely from images or text, both must be considered together. Then, you should output the corresponding chain-of-thought to support your answer. \\

    \textbf{<insert-image-text-pair>} \\

    \textbf{Assistant} \\
    Analysis: <add-analysis-from-model> \\

    \textbf{User} \\
    Unmask the hidden intent!  Given the above rationales, delve into its layers of meaning.  Analyze the surface - the literal words used.  Pierce deeper to uncover the semantic information - the intended meaning behind those words.  Finally, gauge the sentiment - the emotional undercurrent.  By weaving these insights together, can you crack the code of sarcasm and determine if the comment is meant to be sincere or laced with sarcastic? \\
    
    Follow these rules:
    \begin{enumerate}
        \item If any perspective cannot determine sarcasm due to lack of information, disregard that perspective.
        \item If any of the views conflict, choose the view with the most well-founded reasoning.
        \item Output 0 if it does not contain sarcastic content, and 1 if it does. Do not add anything else in your response.
    \end{enumerate}
    
    \end{tcolorbox}
\end{figure}

\end{document}